\documentclass{article}

\usepackage[nonatbib, final]{neurips_2020}

\usepackage[utf8]{inputenc} 
\usepackage[T1]{fontenc}    
\usepackage{hyperref}       
\usepackage{url}            
\usepackage{booktabs}       
\usepackage{amsfonts}       
\usepackage{amsmath}       
\usepackage{nicefrac}       
\usepackage{microtype}      
\usepackage{algorithm}
\usepackage{graphicx}
\usepackage{subcaption}
\usepackage{color}

\usepackage{blindtext}

\usepackage[noend]{algpseudocode} 

\floatname{algorithm}{Algorithm}

\newlength\myindent
\newcommand{\todo}[1]{}
\renewcommand{\todo}[1]{{\color{red} TODO: {#1}}}

\title{Just Pick a Sign: Optimizing Deep Multitask Models with Gradient Sign Dropout}

\author{%
  Zhao Chen\\
  Waymo LLC\\
  Mountain View, CA 94043 \\
  \texttt{zhaoch@waymo.com} \\
  \And
  Jiquan Ngiam \\
  Google Research\\
  Mountain View, CA 94043 \\
  \texttt{jngiam@google.com} \\
  \And
  Yanping Huang \\
  Google Research\\
  Mountain View, CA 94043 \\
  \texttt{huangyp@google.com} \\
  \And
  Thang Luong \\
  Google Research\\
  Mountain View, CA 94043 \\
  \texttt{thangluong@google.com} \\
  \And
  Henrik Kretzschmar \\
  Waymo LLC\\
  Mountain View, CA 94043 \\
  \texttt{kretzschmar@waymo.com} \\
  \And
  Yuning Chai \\
  Waymo LLC\\
  Mountain View, CA 94043 \\
  \texttt{chaiy@waymo.com} \\
  \And
  Dragomir Anguelov \\
  Waymo LLC\\
  Mountain View, CA 94043 \\
  \texttt{dragomir@waymo.com} \\
}

\begin{document}

\maketitle

\begin{abstract}
The vast majority of deep models use multiple gradient signals, typically corresponding to a sum of multiple loss terms, to update a shared set of trainable weights. However, these multiple updates can impede optimal training by pulling the model in conflicting directions. We present Gradient Sign Dropout (GradDrop), a probabilistic masking procedure which samples gradients at an activation layer based on their level of consistency. GradDrop is implemented as a simple deep layer that can be used in any deep net and synergizes with other gradient balancing approaches. We show that GradDrop outperforms the state-of-the-art multiloss methods within traditional multitask and transfer learning settings, and we discuss how GradDrop reveals links between optimal multiloss training and gradient stochasticity. 
  
\end{abstract}

\section{Introduction}

Deep neural networks have fueled many recent advances in the state-of-the-art for high-dimensional nonlinear problems. However, when distilled down to its most basic elements, deep learning relies on the humble \textit{gradient} as the optimization signal which drives its complex algorithmic machinery. Indeed, the desire to properly leverage gradients has spurred a wealth of research into optimization strategies which has led to faster, more stable model training \cite{ruder2016overview}.

However, the literature has habitually glossed over an increasingly crucial detail: most gradient signals are sums of many smaller gradient signals, often corresponding to multiple losses. A broad array of models fall under this category, including ones not traditionally considered multitask; for example, multiclass classifiers can be split into a loss per class, and object detectors conventionally break down their predictions along various bounding box dimensions. It is uncertain, and in fact unlikely, that a na\"ive sum of these individual signals would produce the best solution.

Deep learning theory tells us that the local minima found in single-task models through simple gradient updates are generally of high quality~\cite{choromanska2015loss}. However, such a claim should be reevaluated in the context of multitask loss surfaces, where minima of each constituent loss may exist at different network weight settings, which results in many poor minima of the sum loss. Such undesirable minima are avoided if we encourage the network to seek out critical points that are \textit{joint minima} -- i.e. critical points that lie near a local minimum of all the constituent loss functions. 

To generally address such issues, \textit{deep multitask learning} studies properties of models with multiple outputs and has given birth to methods to balance relative gradient magnitudes \cite{chen2018gradnorm, kendall2018multi} or tune the full gradient tensor \cite{sinha2018gradient}. Still, methods that explicitly tackle joint loss optimization are rare. Works such as \cite{sener2018multi, yu2020gradient} do so by finding a common gradient descent direction for all losses, but such methods operate by \textit{removing} suboptimal gradient components. Such reductive processes are still susceptible to local minima and discourage inter-task competition -- competition which evidence suggests can be beneficial \cite{desideri2012multiple, vandenhende2020revisiting}. Our proposed method not only provides theoretical guarantees of joint loss minima but also allows gradients to compete, and thus avoids the same pitfalls as reductive gradient algorithms. To the best of our knowledge our method is the first with this set of desirable properties.

We motivate our method, Gradient Sign Dropout (GradDrop), by noting that when multiple gradient values try to update the same scalar within a deep network, conflicts arise through differences \textit{in sign} between the gradient values. Following these gradients blindly leads to gradient tug-of-wars and to critical points where constituent gradients can still be large (and thus some tasks perform poorly). 

\begin{figure}[ht]
\centering
\includegraphics[width=0.8\linewidth]{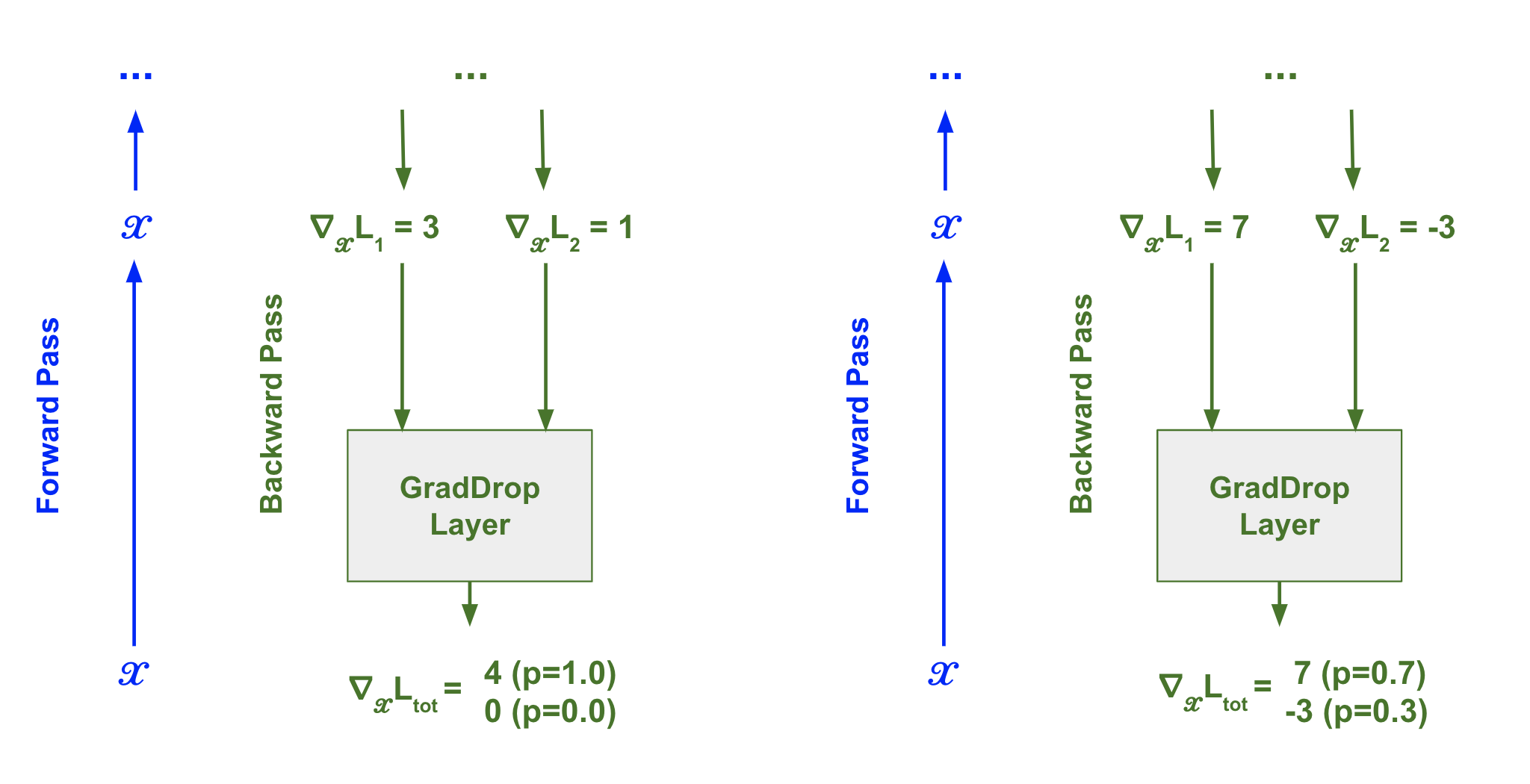}
\caption{GradDrop schematic for two losses and one scalar. In both cases, we calculate $\mathcal{P}$ (from Equation \ref{eq:psp}), which tells us the probability of keeping $\nabla$s with positive signs. On the left, $\mathcal{P} = 0.5*(1 + (3+1)/(|3|+|1|)) = 1.0$, so we keep positive $\nabla$s with 100\% probability. On the right, $\mathcal{P} = 0.5*(1 + (7-3)/(|7|+|-3|)) = 0.7$, so we keep positive $\nabla$s with 70\% probability.}
\label{fig:graddrop}
\end{figure}

To alleviate this issue, we demand that all gradient updates are \textit{pure in sign} at every update position. Given a list of (possibly) conflicting gradient values, we algorithmically select one sign (positive or negative) based on the distribution of gradient values, and mask out all gradient values of the opposite sign. A basic schematic of the method is presented in Figure \ref{fig:graddrop}.

The motivation behind GradDrop parallels the well-known relationship between gradient stochasticity and model robustness \cite{keskar2016large, smith2017don, smith2017bayesian}. When a network finds a narrow, low-quality minimum, the inherent noise within the batched gradient updates serves to kick the model into broader, more robust minima. Similarly, GradDrop assigns a quality score to each gradient update based on its sign consistency, and adds stochasticity along axes where gradients tend to conflict more. An important consequence of this logic is that GradDrop continues triggering until the model finds a minimum that is a joint minimum for all losses (see Section \ref{sec:theory_results} for proof). 



Our primary contributions are as follows:
\begin{enumerate}
    \item We present Gradient Sign Dropout (GradDrop), a modular layer that works in any network with multiple gradient signals and incurs no additional compute at inference.
    \item We show theoretically and in simulation that GradDrop leads to more stable convergence points than na\"ive gradient descent algorithms.
    \item We demonstrate the efficacy of GradDrop on multitask learning, transfer learning, and complex single-task models like 3D object detectors for a variety of network architectures.
\end{enumerate}
\section{Related Work}
\label{section:related_work}

\textbf{Optimization via gradient descent} is one of the key pillars of deep learning. Apart from the traditional optimization methods \cite{duchi2011adaptive, kingma2014adam, nesterov1983method, qian1999momentum, zeiler2012adadelta}, there has been a research thrust on developing different ways to apply gradients to deep networks \cite{chen2018closing, dozat2016incorporating, gradientpenalty, jaderberg2017decoupled, ruder2016overview, tseng2020regularizing, gradientclipping}. The success of such methods comes in part because optimization in single-task models generally converges to high-quality minima \cite{choromanska2015loss}. Also important is the relationship between stochasticity and model robustness; as with GradDrop, noisy gradients help repel poor local minima in favor of wider, more robust critical points \cite{keskar2016large, smith2017don, smith2017bayesian}. These insights are crucial and worth revisiting for multitask environments.

\textbf{Multitask learning} presents a challenging problem for optimization, as the loss surface now consists of many smaller loss surfaces. As a subject of study, multitask learning predates deep learning \cite{caruana1997multitask, desideri2012multiple}, but its power in helping model generalization and transferring information between correlated tasks \cite{meyerson2018pseudo,zamir2018taskonomy} make it especially relevant in the deep learning era. Although a large part of multitask research focuses on developing new network architectures \cite{kaiser2017one, kokkinos2017ubernet, liu2019end, liu2019multi, luong2015multi, ma2018modeling, misra2016cross} or new loss functions \cite{kendall2018multi}, we focus on methods that explicitly interact with the gradients, which tend to be more lightweight and modular. GradNorm \cite{chen2018gradnorm} modifies gradient magnitudes to ensure that tasks train at approximately the same rate. MGDA, the Multiple Gradient Descent Algorithm \cite{desideri2012multiple, sener2018multi}, finds a linear combination of gradients that reduces every loss function simultaneously. PCGrad \cite{yu2020gradient} projects conflicting gradients to each other, which achieves a similar simultaneous descent effect as MGDA. 

\textbf{Many other applications} which are not traditionally considered multitask can benefit from this work. Vision applications such as object detection \cite{liu2016ssd, redmon2016you, ren2015faster, zhou2018voxelnet} and instance segmentation \cite{he2017mask} explicitly construct multiple losses to arrive at one consolidated result. Language models that employ seq2seq predictions \cite{sutskever2014sequence} make multiple predictions and create multiple gradient conflicts when backpropagating through time. Domain adaptation and transfer learning \cite{ganin2014unsupervised, hoffman2017cycada, sun2019unsupervised}, topics in which many powerful specialized techniques have been developed, still often rely on multiple losses and thus can benefit from general multitask approaches. Our approach here, although wrapped in the language of multitask learning, has a much wider range of applicability on deep models in general.
\section{Gradient Dropout}

\subsection{Basic Concepts}

Gradient Sign Dropout is applied as a layer in any standard network forward pass, usually on the final layer before the prediction head to save on compute overhead and maximize benefits during backpropagation. In this section, we develop the GradDrop formalism. Throughout, $\circ$ denotes elementwise multiplication after any necessary tiling operations (if any) are completed.


To implement GradDrop, we first define the Gradient Positive Sign Purity, $\mathcal{P}$, as \begin{equation} \label{eq:psp} \mathcal{P} = \frac{1}{2}\left(1+\frac{\sum_i\nabla L_i}{\sum_i |\nabla L_i|}\right). \end{equation}

$\mathcal{P}$ is bounded by $[0,1]$. For multiple gradient values $\nabla_a L_i$ at some scalar $a$, we see that $\mathcal{P} = 0$ if $\nabla_a L_i< 0 \: \forall i$, while $\mathcal{P} = 1$ if $\nabla_a L_i> 0 \: \forall i$. Thus, $\mathcal{P}$ is a measure of how many positive gradients are present at any given value. We then form a mask for each gradient $\mathcal{M}_i$ as follows: \begin{equation} \label{eq:mask}\mathcal{M}_i = \mathcal{I}[f(\mathcal{P})>U]\circ\mathcal{I}[\nabla L_i >0] + \mathcal{I}[f(\mathcal{P})<U]\circ\mathcal{I}[\nabla L_i <0]\end{equation}
for $\mathcal{I}$ the standard indicator function and $f$ some monotonically increasing function (often just the identity) that maps $[0,1]\mapsto [0,1]$ and is odd around $(0.5,0.5)$. $U$ is a tensor composed of i.i.d $U(0,1)$ random variables. The $\mathcal{M}_i$ is then used to produce a final gradient $\sum \mathcal{M}_i \nabla L_i$.

A simple example of a GradDrop step is given in Figure \ref{fig:graddrop} for the trivial activation $f(x)=x$. 

\subsection{Extension to Transfer Learning and other Batch-Separated Gradient Signals}
\label{sec:batchseparation}

A complication arises when different gradients correspond to different examples, e.g. in mixed-batch transfer learning where transfer and source examples connect to separate losses. The different gradients at an activation layer would then not interact, which makes GradDrop the trivial transformation.

We also cannot just blindly add gradients along the batch dimension, as the information present in each gradient is conditional on that gradient's particular inputs. Generally, deep nets consolidate information across a batch by summing gradient contributions at a trainable weight layer. To correctly extend GradDrop to batch-separated gradients, we will do the same.

For a given layer of activations $A$ of shape $(B, F)$, we imagine there exists an additional weight layer $W^{(A)}$ of shape $(F)$ composed of 1.0s, and consider the forward pass $A \mapsto W^{(A)} \circ A$. $W^{(A)}$ is a virtual layer and is not actually allocated memory during training; we only use it to derive meaningful mathematical properties. Namely, we can then calculate the gradient via the chain rule to arrive at \begin{equation}\label{eq:marginalize} \nabla_{W^{(A)}}L_i = \sum_{\textrm{batch}}(A \circ \nabla_{A}L_i)\end{equation} where the final sum is taken over the batch dimension\footnote{The initialization of the virtual layer is not only meant to keep the forward logic trivial. It is relevant also in the derivation of Equation \ref{eq:marginalize}, as it gives us that $\nabla_AL_i = W^{(A)}\circ \nabla_{W^{(A)}\circ A}L_i = \nabla_{W^{(A)}\circ A}L_i$}. In other words, premultiplying the gradient values by the input allows us to meaningfully sum over the batch dimension to calculate $\mathcal{P}$ and the $\mathcal{M}_i$s. In practice, because we are only interested in $\nabla_{W^{(A)}}$ insofar as it changes the sign content of $\nabla_A$, we will only premultiply by the \textit{sign} of the input.

\subsection{Full GradDrop Algorithm}

The full GradDrop algorithm calculates the sign purity measure $\mathcal{P}$ at every gradient location, and constructs a mask for each gradient signal across $T$ tasks. We specify the details in Algorithm \ref{alg:graddrop}. 

\begin{algorithm}
\caption{Gradient Sign Dropout Layer (GradDrop Layer)}\label{alg:graddrop}
\begin{algorithmic}[1]
\State \textbf{choose} monotonic activation function $f$ \Comment{Usually just $f(p)=p$}
\State \textbf{choose} input layer of activations $A$ \Comment{Usually the last shared layer}
\State \textbf{choose} leak parameters $\{\ell_1, \ldots, \ell_n\} \in [0,1]$ \Comment{For pure GradDrop set all to 0}
\State \textbf{choose} final loss functions $L_1, \ldots, L_n$
\item[]
\Function{BACKWARD}{$A$, $L_1, \ldots, L_n$}\Comment{returns total gradient after GradDrop layer}

\For{$i$ in $\{1, \ldots, n\}$}
\State \textbf{calculate} $G_i = \texttt{sgn}(A)\circ\nabla_A L_i$ \Comment{\texttt{sgn}(A) inspired by Equation \ref{eq:marginalize}}
\If {$G_i$ is batch separated}
\State $G_i \gets \sum_{\texttt{batchdim}} G_i$
\EndIf
\EndFor
\State \textbf{calculate} $\mathcal{P} = \frac{1}{2}\left(1+\frac{\sum_iG_i}{\sum_i |G_i|}\right)$ \Comment{$\mathcal{P}$ has the same shape as $G_1$}
\State \textbf{sample} $U$, a tensor with the same shape as $\mathcal{P}$ and $U[i, j, \ldots] \sim \texttt{Uniform}(0,1)$
\For{$i$ in $\{1, \ldots, n\}$}
\State \textbf{calculate} $\mathcal{M}_i = \mathcal{I}[f(\mathcal{P})>U]\circ \mathcal{I}[G_i >0] + \mathcal{I}[f(\mathcal{P})<U]\circ\mathcal{I}[G_i <0]$
\EndFor
\State \textbf{set} $\textrm{newgrad} = \sum_i (\ell_i + (1-\ell_i)*\mathcal{M}_i)\circ\nabla_AL_i$\\
\Return{newgrad}
\EndFunction
\end{algorithmic}
\end{algorithm}

For many of our experiments, we renormalize the final gradients so that $||\nabla||_2$ remains constant throughout the GradDrop process. Although not practically required, this ensures that GradDrop does not alter the global learning rate and thus observed benefits result purely from GradDrop masking.

Note also the introduction of the leak parameters $\ell_i$. Setting $\ell_i>0$ allows some original gradient to leak through, which is useful when losses have different priorities -- for example, in transfer learning, we prioritize performance on the transfer set. For more details see Section \ref{sec:transferlearning}.

\subsection{GradDrop Theoretical Properties}

We now present and prove the main theoretical properties for our proposed GradDrop algorithm.

\textbf{Proposition 1 (GradDrop stable points are joint minima)}: Given loss functions $L_1, \ldots, L_n$ and any collection of scalars $\bold{W}$ for which $\nabla_{w}L_1, \ldots, \nabla_{w}L_n$ are well-defined, the GradDrop update signal $\nabla^{(GD)}_{w}$ at any position $w\in \mathbf{W}$ is always zero if and only if $\nabla_{w}L_i = 0, \forall i$. 

\textbf{Proof}: Consider $n$ loss functions, indexed $L_1, \ldots, L_n$, and their gradients $\nabla_{w}L_i$ for $w\in \bold{W}$. Clearly, if $\nabla_w L_i = 0, \forall i$, then that $w$ is trivially a critical point for the sum loss $\sum_i L_i$. However, the converse is also true under GradDrop updates. Namely, if there exists some $j$ for which $\nabla_w L_j \neq 0$, without loss of generality assume that $\nabla_w L_j > 0$. According to Equation \ref{eq:psp}, $\mathcal{P} > 0$ at $w$. Thus $f(\mathcal{P})>0$ (as it is monotonically increasing), so there is a nonzero ($f(\mathcal{P})$) chance that we keep all positive signed gradients and thus a nonzero chance that $\nabla^{(GD)}_w \geq \nabla_w L_j > 0$. $\square$

\textbf{Proposition 2 (GradDrop $\nabla$ norms sensitive to \textit{every} loss)}: Given continuous component loss functions $L_i(\mathbf{w})$ with local minima $\mathbf{w^{(i)}}$ and a GradDrop update $\nabla^{(GD)}$, then to second order around each $\mathbf{w^{(i)}}$, $E[|\nabla^{(GD)}L|_2]$ is monotonically increasing w.r.t. $|\mathbf{w}-\mathbf{w^{(i)}}|, \forall i$.

\textbf{Proof}: Set $\boldsymbol\delta := d \boldsymbol \delta_0$ for $|\boldsymbol\delta_0| = 1$. To second order, around a minimum value $\mathbf{w^{(i)}}$ a loss function has the form $L_i(\mathbf{w^{(i)}} + \boldsymbol\delta) \approx L_i(w^{(i)}) + \frac{1}{2}\boldsymbol\delta^TH^{(L_i)}(w^{(i)})\boldsymbol\delta = L_i(w^{(i)}) + \frac{1}{2}d^2\boldsymbol\delta_0^TH^{(L_i)}(w^{(i)})\boldsymbol\delta_0$ 
for positive definite Hessian $H^{(L_i)}$. Because $\boldsymbol\delta_0^TH^{(L_i)}(w^{(i)})\boldsymbol\delta_0>0$, $\nabla L_i$ at $\mathbf{w^{(i)}} + \boldsymbol\delta$ is proportional to $d$. As $d$ increases, so will the magnitude of each $\nabla L_i$ component, which then immediately increases the total expected gradient magnitude induced by GradDrop. $\square$

From Proposition 1, we see that GradDrop will result in a zero gradient update only when the system finds a perfect joint minimum between all component losses. Not only that, but Proposition 2 implies that GradDrop induces proportionally larger gradient updates with distance from \textbf{any} component loss function minimum, regardless of the value of the total loss. The error signals induced by GradDrop are thus sensitive to \textbf{every task}, rather only relying on a sum signal. This sensitivity also increases monotonically with distance from any close local minimum for any component task. Thus, GradDrop optimization will seek out joint minima, but even when such minima do not strictly exist Proposition 2 shows GradDrop will seek out system states that are at least close to joint minima. For a clear illustration of this effect in one dimension, please refer to Section \ref{sec:theory_results}.

A potential concern could be that by being sensitive to every loss function, GradDrop updates are too noisy and the overall system trains more slowly. However, that is not the case, as GradDrop updates \textit{on expectation} are equivalent with standard SGD updates.

\textbf{Proposition 3 (Statistical Properties):} Suppose for 1D loss function $L = \sum_iL_i(w)$ an SGD gradient update with learning rate $\lambda$ changes total loss by the linear estimate $\Delta L^{(SGD)} = -\lambda |\nabla L|^2 \leq 0$. For GradDrop with activation function (see Eq. \ref{eq:mask}) $f(p) = k(p-0.5)+0.5$ for $k\in [0,1]$ (with default setting is $k=1$), we have:
\begin{enumerate}
\itemsep-0.2em
    \item For $k=1$, $\Delta L^{(SGD)} = E[\Delta L^{(GD)}]$
    \item $E[\Delta L^{(GD)}]\leq 0$ and has magnitude monotonically increasing with $k$.
    \item $\text{Var}[\Delta L^{(GD)}]$ is monotonically decreasing with respect to $k$.
\end{enumerate}
We present the proof of this proposition in Appendix \ref{sec:prop3-theory}, along with generalizing it to arbitrary activation functions. $\square$

Importantly, even though GradDrop has a stochastic element, it provides the same expected movement in the total loss function as in vanilla SGD. Also important is the hyperparameter $k$, which controls the tradeoff between how much the GradDrop update follows the overall gradient and how much noise GradDrop induces for inconsistent gradients. A smaller value of $k$ implies a larger penalty/noise scale, and a value of $k=0$ means we randomly choose a sign for every gradient value. We call the $k=0$ case Random GradDrop and show it generally compares unfavorably to $k>0$, but our evidence does not preclude a situation where the higher noise in the $k=0$ case may be desirable. Indeed, in most of our experiments the $k=0$ Random GradDrop setting still outperforms the baseline. 
\section{Experiments with GradDrop}
In this section we present the main experimental results related to GradDrop. All experiments are run on NVIDIA V100 GPU hardware. We will provide relevant hyperparameters within the main text, but we relegate a complete listing of hyperparameters to the Appendix. We also rely exclusively on standard public datasets, and thus move discussion of most dataset properties to the Appendices. 

All multitask baselines (including PCGrad, to keep compute overhead tractable) and the GradDrop layer are applied to the final layer before the prediction heads to keep compute overhead tractable. We primarily compare to other state-of-the-art multitask methods, which include GradNorm \cite{chen2018gradnorm}, MGDA \cite{sener2018multi}, and PCGrad \cite{yu2020gradient}. Descriptions of all these methods were given in Section \ref{section:related_work}. 

For completion, we also compare to Gradient Clipping (e.g. \cite{gradientclipping}) and Gradient Penalty \cite{gradientpenalty}. Although not strictly multitask methods, these gradient-based methods enjoy wide popularity and will provide evidence that principled single-task methods are not enough to optimize a true multitask model.
\subsection{A Simple One-Dimensional Example}
\label{sec:theory_results}

\begin{figure}[ht]
\begin{subfigure}{.33\textwidth}
  \centering
  \includegraphics[width=1.0\linewidth]{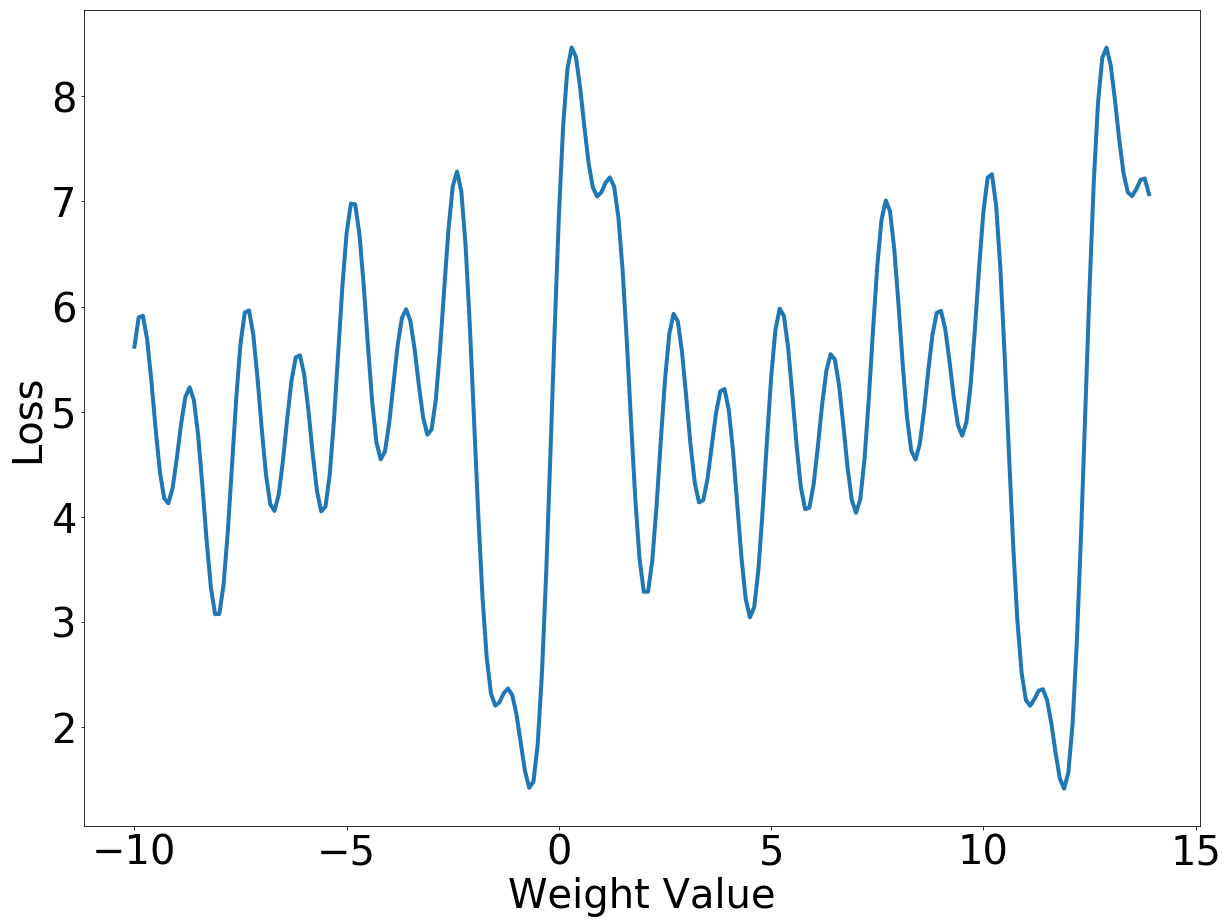}
  \label{fig:theory-sine}
  \caption{Sum of sinusoids loss function}
\end{subfigure}
\begin{subfigure}{.33\textwidth}
  \centering
  \includegraphics[width=1.0\linewidth]{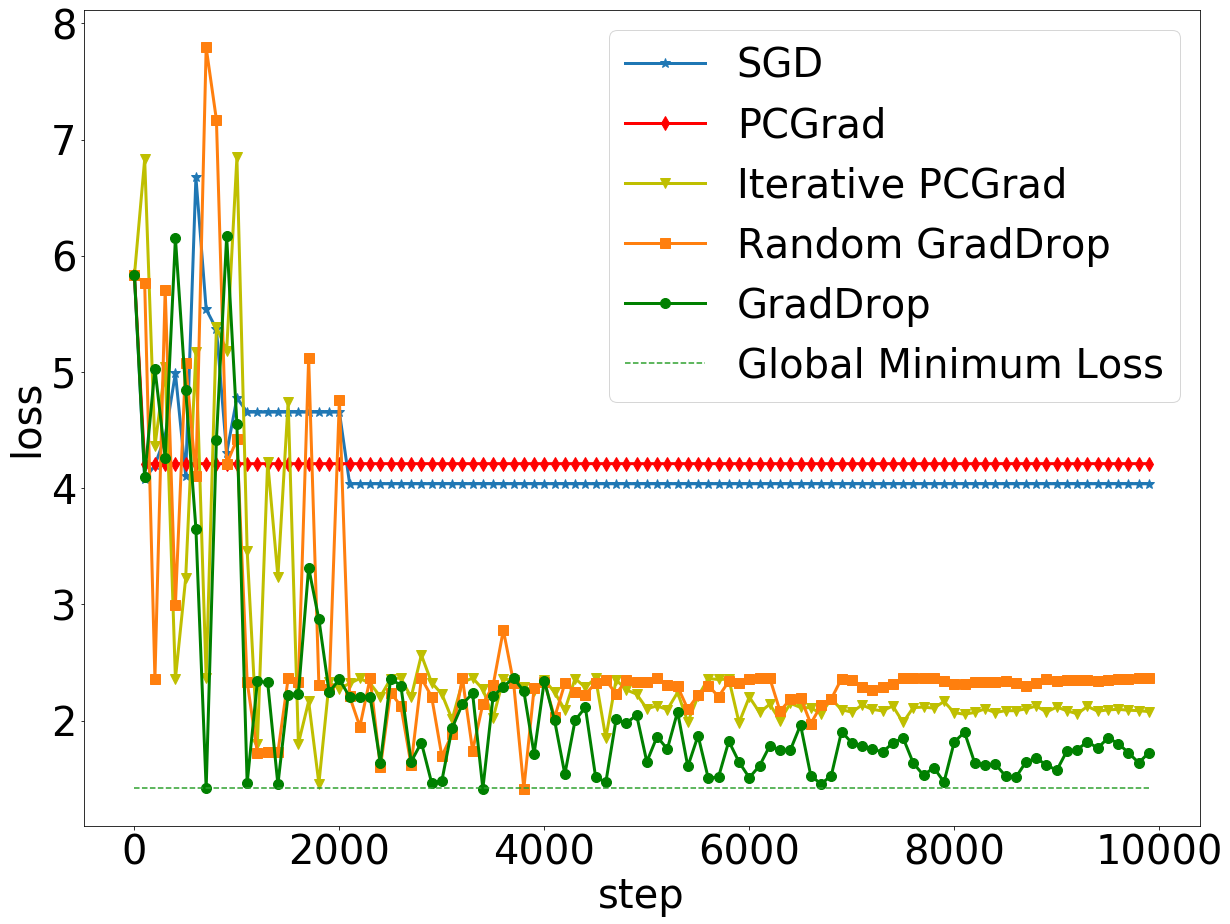}
  \label{fig:theory-singlerun}
  \caption{Loss curves for one random run}
\end{subfigure}
\begin{subfigure}{0.33\textwidth}
  \centering
  \includegraphics[width=1.0\linewidth]{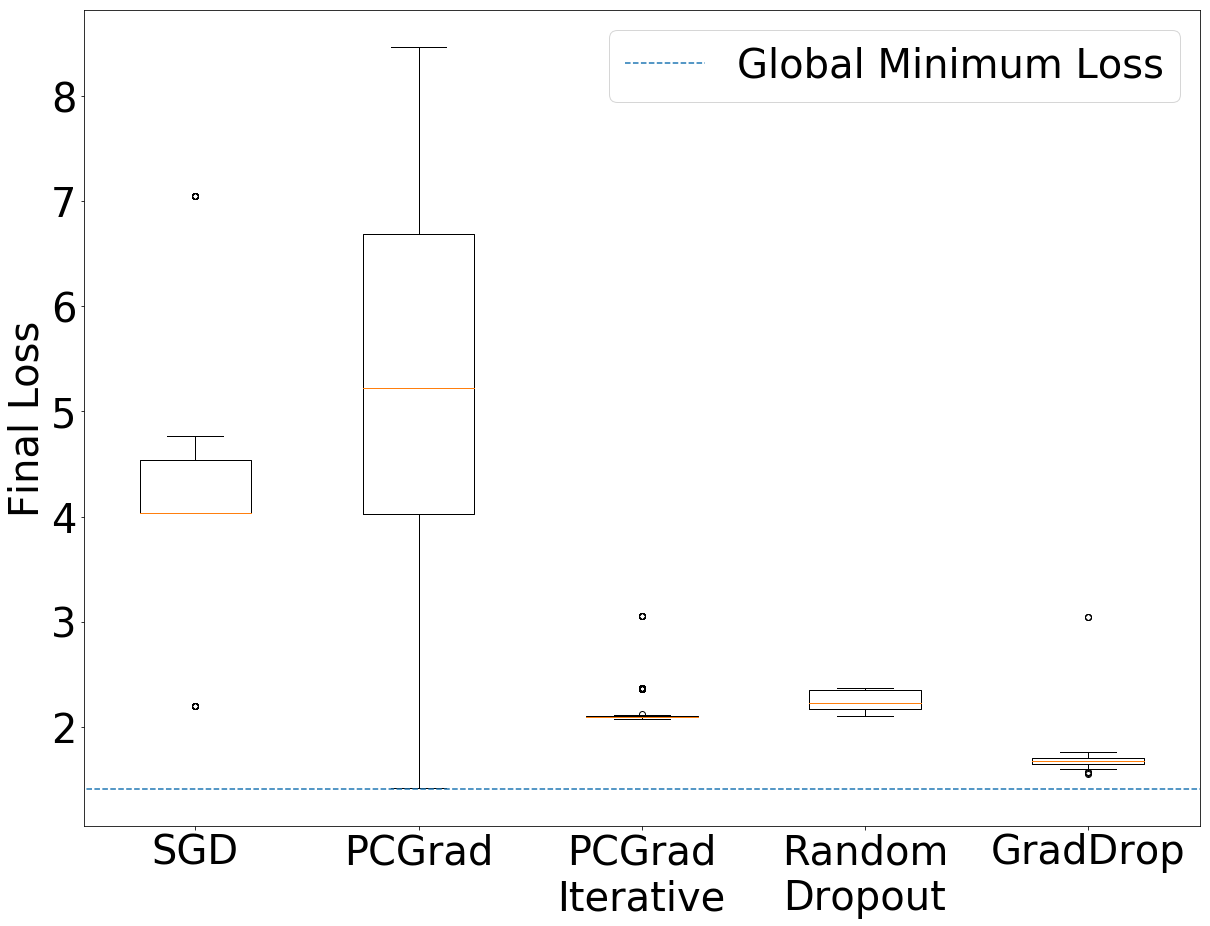} 
  \label{fig:theory-boxplot}
  \caption{Summary results for 200 runs}
\end{subfigure}
\caption{GradDrop toy example. (a) A synthetic 1D loss function composed of five sines. (b) Loss curves for GradDrop and baselines given a random initialization of the trainable weight. (c) Boxplot of final converged loss values when the methods in b. are run 200 times.}
\label{fig:theory}
\end{figure}

We illustrate GradDrop in one dimension. In Figure \ref{fig:theory} we present results on a simple toy system, with a loss function that is the sum of five sines of the form $L(x;a,b) = \sin(ax+b)+1$. The final loss is shown in Figure \ref{fig:theory}(a). Note that although each $L_i$ has identical periodic local minima, the sum loss has a wide distribution of local minima of variable quality.

We now initialize the one weight $w$ to a random value and run various optimization techniques for 10000 steps. In Figure \ref{fig:theory}(b) we plot the loss curves for one example trial. We note that PCGrad \cite{yu2020gradient} does not train in this low-dimensional setting, as any sign conflict would result in PCGrad zeroing the gradients. For fairness, we include a slight modification of PCGrad called iterative PCGrad which still works in low dimensions (for details see Appendix).  We also include Random GradDrop, which is a weak version of GradDrop where $f(\mathcal{P})$ is set to $0.5$ everywhere. We see that GradDrop has the best performance of all methods tested. Such a conclusion is further reinforced when we run this experiment 200 times and plot the statistics of the final results, which are shown in Figure \ref{fig:theory}(c). 

Multiple algorithms (GradDrop, Random GradDrop, and Iterative PCGrad) tend to find the deepest minimum, but GradDrop still performs better. We attribute this to the success of our sign purity measure $\mathcal{P}$ at properly emphasizing gradient directions with higher levels of consistency.

\subsection{Multitask Learning on Celeb-A}
\label{sec:celeba}

We first test GradDrop on the multitask learning dataset CelebA \cite{liu2018large}, which provides 40 binary attributes based on celebrity facial photos. CelebA allows us to test GradDrop in a truly archetypal multitask setting.

\begin{figure}[ht]
\begin{subfigure}{.33\textwidth}
  \centering
  \includegraphics[width=0.92\linewidth]{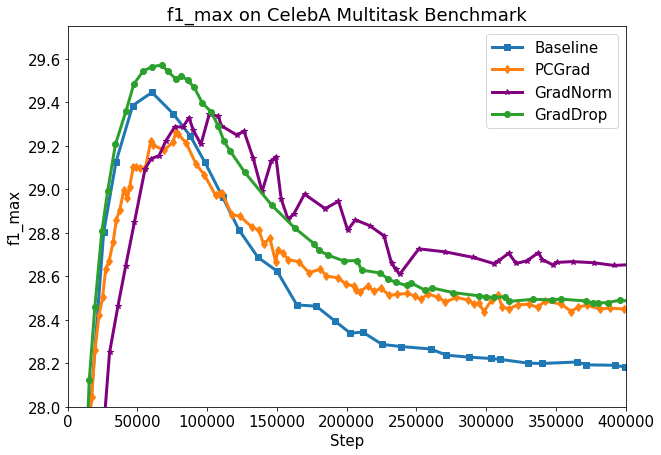}
  \caption{CelebA maximum F1 scores}
\end{subfigure}
\begin{subfigure}{.33\textwidth}
  \centering
  \includegraphics[width=0.92\linewidth]{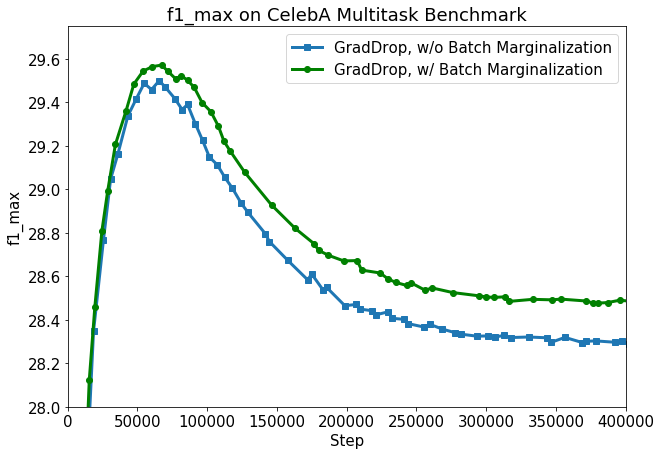}
  \caption{GradDrop batch marginalization}
\end{subfigure}
\begin{subfigure}{.33\textwidth}
  \centering
  \includegraphics[width=0.92\linewidth]{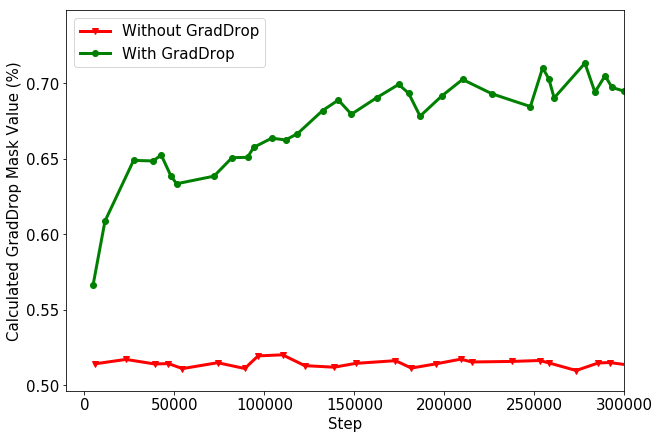}
  \caption{Gradient consistency over time}
\end{subfigure}
\caption{Experiments with GradDrop on CelebA.}
\label{fig:celeba}
\end{figure}

We also use a standard shallow convolutional network to perform this task. Our network consists only of common layers (Conv, Pool, Batchnorm, FC Layers) and contains 9 total layers along with 40 predictive heads. The results of our experiments are summarized in Figure \ref{fig:celeba} and Table \ref{table:celeba}. 

\begin{table}[t]
  \caption{Multitask Learning on CelebA. We repeat training runs and report standard deviations of $\leq 0.04\%$ for F1 Score and $\leq 0.02\%$ for accuracy.}
  \label{table:celeba}
  \centering
  \begin{tabular}{lccc}
    \toprule
    Method     & Error Rate (\%) $\downarrow$    & Max F1 Score $\uparrow$ & Speed Compared to Baseline $\uparrow$ \\
    \midrule
    Baseline     & 8.71 & 29.35 & 1.00     \\
    Gradient Clipping \cite{gradientclipping} &8.70 & 29.34& 1.00\\
    Gradient Penalty \cite{gradientpenalty} &8.63&29.43&0.35\\
    MGDA \cite{sener2018multi}     & 10.82 & 26.00 & 0.25 \\
    PCGrad \cite{yu2020gradient} & 8.72& 29.25& 0.20\\
    GradNorm \cite{chen2018gradnorm} &8.68 &29.32 & 0.41\\
    Random GradDrop &8.60&29.42&\textbf{0.45}\\
    GradDrop (ours) & \textbf{8.52}&\textbf{29.57} & \textbf{0.45}\\
    \bottomrule
  \end{tabular}
\end{table}

We see that GradDrop outperforms all other methods. Although the improvements may seem mild in Table \ref{table:celeba}, they are substantial for this dataset and Figure \ref{fig:celeba}(a) reveals a visually significant effect. Figure \ref{fig:celeba}(b) also shows an ablation study of performance when we choose to marginalize our gradient signal across the batch dimension, as suggested by Section \ref{sec:batchseparation}. Although our gradient signal for CelebA is not batch-separated and thus we are not strictly required to sum the GradDrop signal across our batches, this operation improves GradDrop's memory and compute efficiency, and also can clearly improve model performance. As there are thus few disadvantages from using the sum-over-batch strategy, all further GradDrop runs in this paper will use sum-over-batch.

Furthermore, Figure \ref{fig:celeba}(c) plots the percentage of gradients passed by the GradDrop layer, for both a GradDrop model and a baseline model\footnote{For the baseline model, this statistic is hypothetical and no gradients are actually masked.}. This percentage correlates to the degree of sign consistency of gradients at the GradDrop layer. This metric does not improve at all when training the baseline, but improves appreciably when GradDrop is enabled, suggesting that the critical points found by GradDrop have more consistent gradients and thus higher probability of being a joint minimum.

It is interesting to note that GradDrop also overfits less. We posit that GradDrop is a good regularizer due to its tendency to reject weak loss minima that may overfit. The only stronger regularizer may be GradNorm \cite{chen2018gradnorm}, but GradNorm explicitly curtails overfitting with its $\alpha$ hyperparameter. 

CelebA with its $T=40$ tasks also presents us with an excellent opportunity to test method speed. Looking at the last column of Table \ref{table:celeba}, we see that GradDrop is the fastest of the multitask methods tried (not counting gradient clipping, which is a general single-task method), possibly because it only requires a simple calculation at each tensor position of $\mathcal{O}(T)$ rather than multiple iterative steps like MGDA or $\mathcal{O}(T^2)$ orthogonal projections like PCGrad.
\subsection{Transfer Learning on CIFAR-100}
\label{sec:transferlearning}
We now use GradDrop in a transfer learning setting, which is a batch-separated setting (see Section \ref{sec:batchseparation}). We transfer ImageNet2012 \cite{deng2009imagenet} to CIFAR-100 \cite{krizhevsky2009learning} by using input batches consisting of half CIFAR-100 and half ImageNet2012 examples. Each dataset has its own predictive head and loss.

\begin{figure}[ht]
\begin{subfigure}{.5\textwidth}
  \centering
  \includegraphics[width=0.88\linewidth]{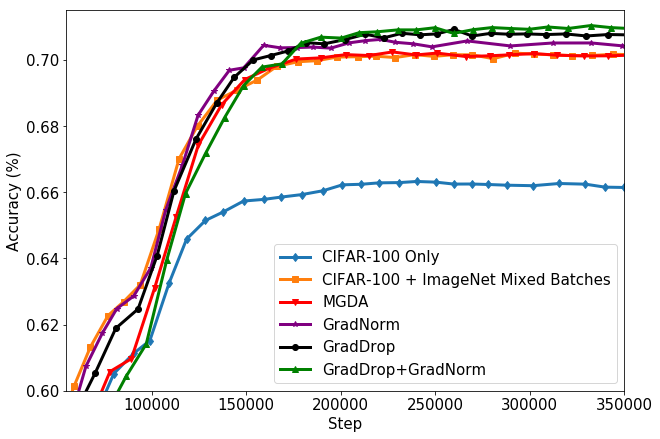}
  \caption{CIFAR-100 accuracy.}
\end{subfigure}
\begin{subfigure}{.5\textwidth}
  \centering
  \includegraphics[width=0.88\linewidth]{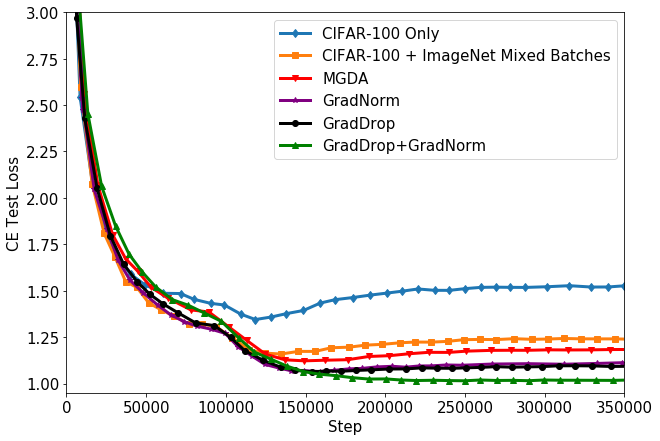}
  \caption{CIFAR-100 loss.}
\end{subfigure}
\caption{Accuracy and loss curves for CIFAR-100 transfer learning experiments. In all cases Gradient Dropout outperforms all other methods tried.}
\label{fig:cifar100transfer}
\end{figure}

We use a more complex network based on DenseNet-100 \cite{huang2017densely}, both to increase performance and to test GradDrop with more complex network topologies. Our results are shown in Table \ref{table:cifar100} and Figure \ref{fig:cifar100transfer}, where we present the best accuracy achieved by each method and the corresponding loss\footnote{This is the loss that corresponds to the highest accuracy model, not the model with the lowest loss. However, reporting the latter would not change the trend.}; we include the loss as it is generally smoother.  

\begin{table}
  \caption{Transfer Learning from ImageNet2012 to CIFAR-100. We repeat training runs and observe standard deviations of $\leq 0.2\%$ accuracy and $\leq 0.01$ loss.}
  \label{table:cifar100}
  \centering
  \begin{tabular}{lcc}
    \toprule
    Method     & Top-1 Error (\%) $\downarrow$ &  Test Loss $\downarrow$ \\
    \midrule
    Train on CIFAR-100 Only & 33.6 & 1.52      \\
    Mixed Batch (MB)     & 29.8   & 1.22    \\
    MB + Gradient Clipping \cite{gradientclipping}& 29.4 & 1.22\\
    MB + Gradient Penalty \cite{gradientpenalty}& 30.6 & 1.28\\
    MB + MGDA \cite{sener2018multi}     & 29.7 & 1.17        \\
    MB + GradNorm \cite{chen2018gradnorm} &29.4 & 1.11\\
    MB + GradDrop (ours) & 29.1 & 1.08\\
    MB + GradNorm \cite{chen2018gradnorm} + Random GradDrop & 29.8 & 1.04\\
    MB + GradNorm \cite{chen2018gradnorm} + GradDrop (ours) & \textbf{28.9}&\textbf{1.01}\\
    \bottomrule
  \end{tabular}
\end{table}

We see that the best model uses a combination of GradDrop and GradNorm \cite{chen2018gradnorm}, although the GradDrop-Only model also performs well. As in the CelebA experiments presented in Section \ref{sec:celeba}, the performance gap is larger when the baseline models overfit later in training. The general synergy between GradDrop and other multitask methods such as GradNorm is important, as it suggests GradNorm can add to complex models which already employ an array of pre-existing deep learning tools. We explore this synergy further in Section \ref{sec:synergy}.

For our final GradDrop model we use a leak parameter $\ell_i$ set to 1.0 for the \textit{source set}. In this setting, source set gradients are allowed to flow unimpeded but transfer set gradients are masked. This setting is optimal as the source dataset is usually larger and the masking effectively curtails overfitting on the transfer dataset. For more experiments related to the leak parameter, see Section \ref{sec:appendix-cifar100}. 
\subsection{3D Point Cloud Detection on Waymo Open Dataset}
We now present results on a much more complex problem: 3D vehicle detection from point clouds on the Waymo Open Dataset \cite{sun2019scalability}. For this task we use a PointPillar model \cite{lang2019pointpillars}, a complex and competitive 3D detection architecture that voxelizes a point cloud and uses standard 2D convolutions to derive deep predictive features. We also note that object detection is traditionally considered a \textit{single-task} problem, but still has multiple losses -- 3 for each coordinate of the box centers, 3 for each dimension of the box, 1 on box orientation, and (in our formulation) 2 classifiers for box motion direction and box class. Our results thus show that GradDrop is applicable in a much wider context than the traditional explicit interpretation of ``multitask learning" might imply. 


\begin{table}
  \caption{Object Detection from Point Clouds on the Waymo Open Dataset. We report standard deviations of $\leq 0.3\%$ on AP values and $\leq 0.5\%$ on APH values.}
  \label{table:waymoopendataset}
  \centering
  \begin{tabular}{lcccc}
    \toprule
    Method     & 2D AP (\%) $\uparrow$    & 2D APH (\%) $\uparrow$& 3D AP (\%)$\uparrow$& 3D APH (\%) $\uparrow$\\
    \midrule
    Baseline &76.2&69.9&57.1 &53   \\
    \midrule
    \textbf{Gradient Norm Methods} \\
    MGDA \cite{sener2018multi}& \textbf{76.8}&69.5&20.0&18.3   \\
    GradNorm \cite{chen2018gradnorm}& \textbf{76.9}&71.7&51.0&48.2   \\
    \midrule
    \textbf{Full Gradient Tensor Methods} \\
    PCGrad \cite{yu2020gradient}& 76.2&70.2&58.4&54.4   \\
    Random GradDrop & 76.4&66.6&57.6&50.5   \\
    GradDrop (Ours) & \textbf{76.8}&\textbf{72.4}&\textbf{58.8}&\textbf{56.0}   \\
    \bottomrule
  \end{tabular}
\end{table}

Our main results are shown in Table \ref{table:waymoopendataset}, where we show Average Precision (AP) and Average Precision w/ Heading (APH) scores (for training curves see Appendix). APH is a metric introduced in \cite{sun2019scalability}, which penalizes boxes for being $180^o$ mis-oriented. All runs include gradient clipping at norm 1.0, and we are unable to compare to gradient penalty due to memory restrictions. GradDrop results in marked improvements, especially in the APH metrics. We also note that like the gradient norm methods (which focus on the overall magnitude of gradients rather than their high-dimensional content), GradDrop provides a moderate boost in 2D performance. However, GradDrop does not suffer from the same substantial regressions in 3D performance, and instead improves all metrics across the board. 
\subsection{Synergy with Gradient Normalization and Other Methods}
\label{sec:synergy}

One important property of GradDrop is that it primarily modifies the gradient tensor direction, which is then largely left alone by other deep learning techniques. In principle, GradDrop can thus be applied in parallel with other multitask methods. In this section, we demonstrate positive interactions between GradDrop an GradNorm \cite{chen2018gradnorm}, evidence that GradDrop can be considered a modular part of a diverse toolset which can be applied in a wide array of applications. 

\begin{table}
  \caption{Synergy Between GradDrop and GradNorm}
  \label{table:celeba_synergy}
  \centering
  \begin{tabular}{lcccc}
  \toprule
  &\multicolumn{2}{c}{\textbf{CelebA}}&\multicolumn{2}{c}{\textbf{Waymo Open Dataset}}\\
    \midrule
    Method     & Err Rate (\%) $\downarrow$  & F1$_{\textrm{max}}$ $\uparrow$ & 3D AP (\%) $\uparrow$ & 3D APH (\%) $\uparrow$   \\
    \midrule
    GradNorm Only & 8.68 & 29.32 & 51.0 & 48.2\\
    GradNorm + GradDrop (ours) & \textbf{8.57} & \textbf{29.50} &\textbf{55.1}& \textbf{51.5}\\
    \bottomrule
  \end{tabular}
\end{table}


Our main results regarding synergy between GradDrop and GradNorm are summarized in Table \ref{table:celeba_synergy}. Along with the CIFAR-100 results in Section \ref{sec:transferlearning}, we find GradDrop often leads to significant improvements when applied with GradNorm. This is especially true where GradNorm performs poorly; for example, although GradNorm tends to regress in the 3D AP metrics compared to baseline, GradDrop+GradNorm recovers much of that performance while still performing well in the 2D AP metrics (see Appendix for 2D AP numbers). We also experimented with GradDrop+MGDA, but with limited success. We hypothesize that MGDA works best when input tensors have explicitly conflicting signs, while GradDrop's final gradient tensors have the same sign (or zero) at all positions. 

From an efficiency standpoint, applying GradDrop on top of GradNorm or MGDA comes essentially for free; both GradNorm and MGDA already require us to calculate $\nabla_{\mathbf{W}}L_i, \forall i$, which is the most expensive step in GradDrop. And because we know GradDrop is faster than the other methods described (see Table \ref{table:celeba}), the additional compute to add GradDrop is small.

\section{Conclusions}
We have presented Gradient Sign Dropout (GradDrop), a method that turns additive gradient signals into a sum signal that is pure in sign and encourages the network to seek out joint minima. From a theoretical standpoint, GradDrop provides superior behavior in the face of suboptimal local minima, and also works for a wide array of network architectures and multitask learning settings.

Apart from our concrete contributions, we also hope that GradDrop will invigorate discussion regarding how best to optimize the complex loss surfaces induced by multitask learning. Our results suggest that the traditional faith in standard gradient descent methods may not describe the full picture, and a realignment of our understanding of optimization robustness to include multitask concepts and gradient stochasticity is prudent as models become ever more complex. We present GradDrop as a crucial early piece of this increasingly important puzzle.
\newpage
\section{Broader Impacts}
In this paper we presented GradDrop, a general algorithm that can be used as a modular addition to multitask models. At its core, our contribution is the development of a general machine learning algorithm without any assumptions of specific applications, so the potential broader impacts of our work is dependent on the application area. 

However, it is also true that multitask learning operates by attempting to leverage multiple sources of potentially disparate information and making joint predictions based on those sources. When applied correctly, multitask models can be less prone to bias/unfairness as they have access to a larger, more diverse source of information. However, when applied incorrectly, multitask models may end up reinforcing the same biases that we want to eliminate; imagine, for example, multitask models which make predictions separately for different subpopulations of the input dataset and due to lack of proper training dynamics end up overfitting to each in turn. Our proposed algorithm may have beneficial effects in combating such overfitting, as our algorithm is effective at finding joint solutions that consistently take all available information into account. As such, we believe that GradDrop will have a positive broader impact on machine learning work by providing ways to arrive at better regularized solutions that are more reflective of reality.
{\small
\bibliographystyle{ieee}
\bibliography{egbib}
}

\newpage
\appendix

\section{Appendix}

The majority of the appendix is devoted to a faithful listing of hyperparameters, datasets, and training settings for all of our experiments. However, we also expand on some intuitions behind our treatment of batch-separated gradients in Section \ref{sec:appendix-batch-separation} and present some more experiments on CIFAR-100 transfer learning in Section \ref{sec:appendix-cifar100} and on the Waymo Open Dataset in Section \ref{sec:appendix-wod}.

\subsection{Addendum on Proposition 3 and Choice of Activation Function}\label{sec:prop3-theory}

We begin with a proof of Proposition 3, which we rewrite here for convenience:

\textbf{Proposition 3:} Suppose for 1D loss function $L = \sum_iL_i(w)$ an SGD gradient update with learning rate $\lambda$ changes total loss by the linear estimate $\Delta L^{(SGD)} = -\lambda |\nabla L|^2 \leq 0$. For GradDrop with activation function $f(p) = k(p-0.5)+0.5$ for $k\in [0,1]$ (with default setting is $k=1$), we have:
\begin{enumerate}
\itemsep-0.2em
    \item For $k=1$, $\Delta L^{(SGD)} = E[\Delta L^{(GD)}]$
    \item $E[\Delta L^{(GD)}]\leq 0$ and has magnitude monotonically increasing with $k$.
    \item $\text{Var}[\Delta L^{(GD)}]$ is monotonically decreasing with respect to $k$.
\end{enumerate}

\textbf{Proof:} For simplicity of notation and without loss of generality, let us assume a learning rate of $\lambda = 1$. Define $p:=\sum_{\nabla_i \geq 0} |\nabla_i|$ and $n:=\sum_{\nabla_i < 0} |\nabla_i|$ to be the total absolute value of positive and negative gradients, respectively. From the definition of $\mathcal{P}$ as in Eq. \ref{eq:psp}, we can easily derive that $\mathcal{P} = p/(p+n)$. 

We then calculate \begin{equation} f(\mathcal{P}) = k(\mathcal{P} - 0.5) + 0.5  = 0.5\left(\frac{p-n}{p+n}\right)k + 0.5\end{equation}
\begin{equation} 1-f(\mathcal{P}) = -0.5\left(\frac{p-n}{p+n}\right)k + 0.5\end{equation}

We then note that with total gradient $p-n$, the value $\Delta L$ under GradDrop is precisely \begin{equation} E[\Delta L^{(GD)}] = -(p-n)(f(\mathcal{P})p + (1-\mathcal{P})(-n))\end{equation}
\begin{equation} = -(p-n)\left(0.5\left(\frac{p-n}{p+n}\right)kp + 0.5\left(\frac{p-n}{p+n}\right)kn + 0.5p - 0.5n\right)\end{equation}
\begin{equation} = -0.5(p-n)\left(\left(\frac{k}{p+n}\right)((p-n)p+(p-n)n)+(p-n)\right)\end{equation}
\begin{equation} = -0.5(p-n)\left(k(p-n)+(p-n)\right)\end{equation}
\begin{equation} = -0.5(k+1)(p-n)^2\end{equation}

We note that for $k=1$, this reduces to $-(p-n)^2$, which is precisely $\Delta L^{(SGD)}$, proving the first claim. We also note that the magnitude of this expression is monotonically increasing with $k$, but it is always negative assuming $k\geq 0$, thus proving the second claim. 

As for the variance claim, it is straightforward to calculate:

\begin{equation} \text{Var}[\Delta L^{(GD)}] = E[(\Delta L^{(GD)})^2] - (E[(\Delta L^{(GD)}])^2\end{equation}

\begin{equation}  = (f(\mathcal{P})p^2 + (1-f(\mathcal{P}))n^2)(p-n)^2 - (0.5(k+1)(p-n)^2)^2\end{equation}

\begin{equation} = 0.5(p-n)^2\left(\left(\frac{p-n}{p+n}\right)p^2k - \left(\frac{p-n}{p+n}\right)n^2k + p^2 + n^2 \right)- (0.5(k+1)(p-n)^2)^2\end{equation}

\begin{equation} = 0.5(p-n)^2\left(\left(\frac{p-n}{p+n}\right)(p^2-n^2)k + p^2 + n^2 - 0.5(k+1)^2(p-n)^2 \right)\end{equation}
\begin{equation} = 0.5(p-n)^2\left((p-n)^2k + p^2 + n^2 - 0.5(k+1)^2(p-n)^2 \right)\end{equation}
\begin{equation} = 0.5(p-n)^2\left((p-n)^2(k-0.5(k+1)^2) + p^2 + n^2 \right)\end{equation}
\begin{equation} = 0.25(p-n)^2\left((p-n)^2(-k^2 - 1) + 2p^2 + 2n^2 \right)\end{equation}

Although not as simple as our expression for expected value, the variance expression treated as a function of $k$ looks like $A(-k^2-1) + B$, with $A, B\geq 0$ and is thus clearly a monotonically decreasing function of $k$ for $k\in [0,1]$. The third claim is proven. $\square$

Although Proposition 3 was proven for a specific family of activation functions (i.e. $f(p) = k(p-0.5) + 0.5$), it easily extends to the result that any choice of $f$ that is (1) odd around the point $(0.5, 0.5)$, (2) monotonically increasing, and (3) bounded by $0.0 \leq f(p)\leq 1.0$ will have similar characteristics. Namely, the \textit{steeper} (formal definition to follow) that $f$ is, the higher its corresponding magnitude of $E[\Delta L^{(GD)}]$ and the lower its variance. Namely,

\textbf{Corollary 3.1}: Take the family of real-valued continuous activation functions $\mathcal{F}$ such that $f\in \mathcal{F}$ if $f$ is defined on the domain $[0,1]$, odd around $(0.5, 0.5)$, monotonically increasing, and has output bounded by $0\leq f(p)\leq 1$ on its domain. We say $f\in \mathcal{F}$ is steeper than $g$ if $f(p)\geq g(p)$ when $p\geq 0.5$ and $f(p)\leq g(p)$ otherwise. For $f, g\in \mathcal{F}$, if $f$ is steeper than $g$, call the corresponding expected loss changes as $E[\Delta L^{(f)}]$ and $E[\Delta L^{(g)}]$. Then the following must be true: 

\begin{enumerate}
\itemsep-0.1em
    \item $E[\Delta L^{(f)}] \leq E[\Delta L^{(g)}] \leq 0$.
    \item $\text{Var}[\Delta L^{(f)}] \leq \text{Var}[\Delta L^{(g)}]$.
\end{enumerate}

\textbf{Proof}: It is important to note that the proof for Proposition 3 is true for all values of $p\geq 0$ and $n\geq 0$. That is, the proof for Proposition 3 immediately implies that given any triplet of values ($p, n, \mathcal{P}$), the claims of the proposition are true as a function of $k$. For $\mathcal{P} = 1$, tuning the value of $k$ allows us to sweep the value of $f(1)$ smoothly from 0.5 to 1, and the corresponding value of $f(-1)$ smoothly from 0.5 to 0. Thus, at these two special points, we have access to the full range of possible outcomes. And so if we limit ourselves to the $\mathcal{P} = 1$ and $\mathcal{P} = 0$ cases, we immediately conclude the following:

Given any value of $(p, n)$ and the resultant value of $\mathcal{P}$,  if $f(\mathcal{P}) \geq g(\mathcal{P})$ and $\mathcal{P} \geq 0.5$, or if $f(\mathcal{P}) \leq g(\mathcal{P})$ and $\mathcal{P} \leq 0.5$, then $E[\Delta L^{(f)}] \leq E[\Delta L^{(g)}]$ and $\text{Var}[\Delta L^{(f)}] \leq \text{Var}[\Delta L^{(g)}]$ as a special case of Proposition 3. 

Because the conditions so listed cover every value for every possible valid  activation function $f$ and $g$, the corollary is proven. We also note that $E[\Delta L^{(f)}] \leq 0$ for any $f\in \mathcal{F}$ because the ``least steep" activation function is $f(p) = 0.5$, which we showed in Proposition 3 has an expected $\Delta L^{(f)}$ value of $\leq 0$. $\square$

We note that the variance claims in both Proposition 3 and Corollary 3.1 are relatively simple extensions of the intuitive result that the variance of a random variable that can take on only two values is maximized when the two values each have a probability weight of $50\%$. We also note that because of the corrollary, the results in Proposition 3 are in fact valid for the extended class of activation functions $f(p) = \text{clip}(k(p-0.5)+0.5, 0.0, 1.0)$ for all $k\geq 0$.

\subsection{More Intuition Regarding Batch-Separated Gradients}\label{sec:appendix-batch-separation}

Perhaps one of the most subtle components of the proposed GradDrop method is its treatment of batch-separated gradients. Although the treatment in the main paper is more mathematical, we would like to use this section to develop some more intuition for our proposed methodology. 

As described in Section \ref{sec:batchseparation}, it is necessary to develop a version of GradDrop that operates nontrivially when gradients are incident on orthogonal sub-batches, like in our transfer learning experiments in Section \ref{sec:transferlearning}. The issue we need to resolve is that these gradients are dependent on their batch's input values, so just summing gradients across the batch dimension is not an option. For example, a gradient value of 4.0 when the input value is 1.0 is not in general the same scenario as a gradient value of 4.0 when the input value is -1.0. 

An important insight is that most operations in a standard deep network are \textit{multiplicative} in nature. Although additions of a bias are also standard in neural networks, they are vastly outnumbered by the amount of multiplicative operations and often are left out entirely of the network. However, if our basic building block within a deep network is multiplication, this means that the important quantity is not the pure value of a gradient, but whether that gradient pulls an input value \textit{further or closer to zero}. Thus, the important value when comparing gradients is (input)$\times$(grad), rather than the naked gradient. 

However, an additional complication arises because the input is often high variance, and taking this product as our key metric can produce unstable results. An additional modification can be made based on the reasoning that GradDrop operates mainly by reasoning about the sign content of the gradients. The reason why pre-multiplication by the input value is useful is only because it ensures we do not make a sign error when summing multiple gradients together. In that sense, it is sufficient to premultiply by the \textit{sign} of the input, as this allows us to correct our gradient signal for any potential sign errors without being susceptible to the added variance of the inputs. 

In the main paper Section \ref{sec:batchseparation}, we derived the proposed rule by assuming a virtual layer that was simple element-wise multiplication at each activation position. In principle, there are also other layers with trainable weights (e.g. dense layers, conv layers) for which we could consider a virtual layer and derive a rule for marginalization of the gradient signal across batches. It is a potential direction of future work to see if any of these other layers result in more robust rules for gradient comparison. 

\subsection{A Simple One-Dimensional Example: Addendum}

Because the experiment in Section \ref{sec:theory_results} uses a model with only one trainable weight, there isn't much to list in terms of hyperparameters. We train all runs with an initial learning rate of 0.2 and a decay ratio of 0.5 applied every 1k steps. Every run is 10k steps in total. We use a standard SGD optimizer.

The sine curves we use to generate the final loss are of the form $\sin(ax+b)+1.0$. The 1.0 affine factor is only there so that all loss values are nonnegative, which is purely cosmetic. The five sine functions have the following parameters for $(a,b)$: 

\begin{center}
(1.0, 0.0)\\
(1.5, 0.2)\\
(2.0, 0.4)\\
(2.5, 0.6)\\
(5.0, 0.8)
\end{center}

The sine function periods are selected purely at random and are not chosen to necessarily emphasize any particular behavior.

We note that many methods, such as MGDA \cite{sener2018multi} and PCGrad \cite{yu2020gradient} do not operate well in the low-dimensional regime. Although it is difficult to adapt MGDA to lower dimensions, we were able to modify PCGrad to exhibit nontrivial behavior in low dimensions. Namely, PCGrad first makes a static copy of the original gradients and then orthogonally projects gradient tensors to each other with reference to the static copy. Instead, we do not make a copy of the original gradient vector and instead update the input gradients in-place. Such a replacement strategy, which we call Iterative PCGrad, adds noise to the PCGrad method but allows for reasonable operation in low dimensions. We also have tried Iterative PCGrad on some of the other experimental settings within this work and it generally seems to perform similarly to PCGrad proper. 

\subsection{Multitask Learning on Celeb-A: Addendum}

For these experiments we use the Celeb-A dataset in its standard setting. We use the standard 160k/20k dataset split and treat each attribute as a separate task that is trained with a standard binary sigmoid classification loss.

Our network is a shallow convolutional network with nine layers (not counting the maxpool layers or predictive head). With the notation CONV-$F$-$C$ for a convolutional layer of filter size $F$ and number of channels $C$, MAX denoting a maxpool layer of filter size and stride 2, and DENSE-H a dense layer with $H$ outputs, the layer stack is [CONV-3-64][MAX][CONV-3-128][CONV-3-128][MAX][CONV-3-256][CONV-3-256][MAX][CONV-3-512][CONV-3-512][DENSE-512][DENSE-512][DENSE-40]. GradDrop and other baselines are applied after the final CONV layer. All layers use Batch Normalization \cite{ioffe2015batch} except for the final predictive head. 

We use an Adam optimizer with $(\beta_1,\beta_2)$ = (0.9, 0.999). Our batch size is 8 and we start with a learning rate of 1e-3, with an annealing rate of 0.96 applied every 2400 steps. All baselines are trained with this set of hyperparameters, with the exception of MGDA \cite{sener2018multi} for which we had to lower the learning rate by a factor of 100x (otherwise the performance of MGDA was very poor). We train all networks past convergence, but report results from the performance peak. We do this so we also can see the behavior of the system when the model degradation from overfitting is most pronounced. 

\subsection{Transfer Learning on CIFAR-100: Addendum}
\label{sec:appendix-cifar100}

We use CIFAR-100 in its standard setting with a 40k/10k data split. All images (including the ImageNet2012 images) are resized to 32x32 before being input into the network to match the CIFAR-100 image resolution. Image values are divided by 256.0 in preprocessing so that values input to the network lie between 0 and 1. This initial normalization improves training stability especially at the beginning of training. 

Our network is based on a DenseNet-100-BC \cite{huang2017densely} model with $k=12$. The model has 100 layers in total. We do not use data augmentation to reduce the variance of our training results, and we search for hyperparameters that perform optimally on the transfer learning baseline before applying other baselines with the same set of hyperparameters. We do not use Dropout \cite{srivastava2014dropout} in our network as it appears to degrade performance. We use batch size 8 each for CIFAR-100 and ImageNet2012 inputs (for a total batch size of 16), with an Adam optimizer with $(\beta_1, \beta_2) = (0.9, 0.999)$ and an initial learning rate of 0.001. The learning rate stays constant until step 100k, at which point it decays by a factor of 0.94 every 2000 steps. We train until convergence, which occurs at around 250k ($\approx$50 epochs). Our best results use the GradDrop activation $f(p) = 0.25(p-0.5)+0.5$, showing that this particular system benefits from a higher noise penalty (i.e. see theoretical results in Section \ref{sec:prop3-theory}). 

Both ImageNet2012 and CIFAR-100 inputs share the vast majority of the network, but they are given separate BatchNorm trainable parameters to help alleviate negative effects of the domain shift between the two datasets. We found that training is very unstable in this transfer learning setting without this BatchNorm parameter separation.  

Unlike our other experiments, we do not try PCGrad \cite{yu2020gradient}. This is primarily because PCGrad is the trivial transformation when all gradients have nonnegative pairwise dot products. However, because transfer learning produces batch-separated gradient signals (see Section \ref{sec:batchseparation}), all the gradients are already pairwise orthogonal before any additional processing. Thus PCGrad would return the trivial transformation and would perform identically to the baseline. 

\begin{figure}[ht]
\begin{subfigure}{.5\textwidth}
\captionsetup{width=0.8\textwidth}
  \centering
  \includegraphics[width=\linewidth]{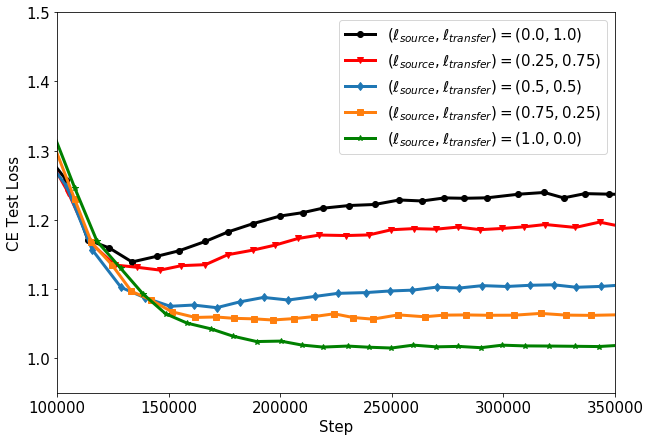}
  \caption{CIFAR-100 transfer learning with various leak parameter settings.}
\end{subfigure}
\begin{subfigure}{.5\textwidth}
\captionsetup{width=0.8\textwidth}
  \centering
  \includegraphics[width=\linewidth]{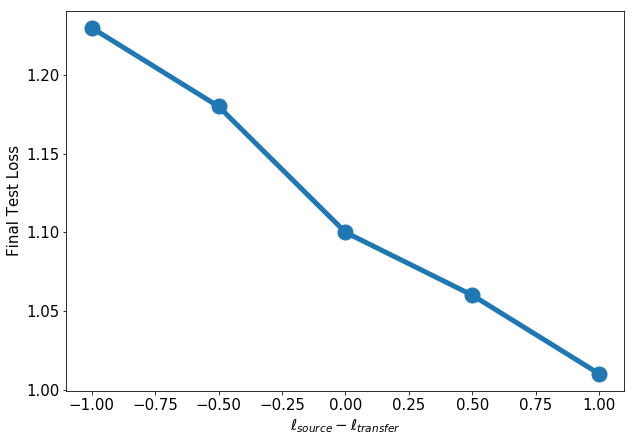}
  \caption{Final CIFAR-100 transfer learning loss plotted against leak parameter settings.}
\end{subfigure}
\caption{Experiments with leak parameters on the CIFAR-100 transfer learning setting.}
\label{fig:cifar_leak}
\end{figure}

\begin{table}
  \caption{Transfer Learning from ImageNet2012 to CIFAR-100 with different leak parameters. Standard deviation values are $\leq 0.2\%$ for accuracy and $\leq 0.01$ for loss.}
  \label{table:cifar100-leak}
  \centering
  \begin{tabular}{cccc}
    \toprule
    $\ell_{source}$ & $\ell_{transfer}$     & Top-1 Error (\%) $\downarrow$ &  Test Loss $\downarrow$ \\
    \midrule
    0.0 & 1.0 & 30.5 & 1.23      \\
    0.25 & 0.75 & 30.0   & 1.18    \\
    0.5 & 0.5   & 29.0 & 1.10        \\
    0.75 & 0.25 &29.1 & 1.06\\
    1.0 & 0.0 & \textbf{28.9} & \textbf{1.01}\\
    \bottomrule
  \end{tabular}
\end{table}

In the main paper, we also noted that GradDrop allows for flexibility in leak parameters $\ell_i \in [0.0, 1.0]$, such that the final gradient returned is $\ell_i\nabla + (1-\ell_i)\nabla^{(\textrm{graddrop})}$ for a given task $i$. We made the claim that for a transfer learning setting, having the standard $\ell_i = 0, \forall i$ environment is suboptimal as we care more about performance on the transfer task. We further claimed that setting a leak parameter of $\ell=1.0$ for the \textit{source} dataset while keeping $\ell=0.0$ for the transfer dataset was the optimal setting for transfer learning. 

We present here experiments that empirically justify the above statement. In Table \ref{table:cifar100-leak} and Figure \ref{fig:cifar_leak} we show results of a panel of experiments conducted with different leak parameters $\ell_{source}$ and $\ell_{transfer}$. We run the CIFAR-100 experiment with five different settings of $\ell_{source}$ and $\ell_{transfer}$, although for ease of interpretation we keep the sum $\ell_{source} + \ell_{transfer}$ at a constant value of 1.0. We note that there is a clear dependence of performance on the value $\ell_{source}-\ell_{transfer}$. The error values stay generally the same for $\ell_{source}$ close to 1.0, but then rise precipitously, although the same trend manifests as a strong linear dependency in the loss values. 

To some, this result may be counterintuitive; if we care more about the transfer set, then it seems reasonable that $\ell_{transfer}$ should be higher and not $\ell_{source}$, to ensure that more transfer gradients are transmitted back through the network. However, we find these results fully consistent with our understanding of GradDrop; as GradDrop primarily filters for consistent gradients, it is optimal to allow the unimportant source set to fully overfit while the transfer set is maximally filtered and regularized. We find that this set of experiments strongly suggests that the effect of GradDrop is beneficial. 

\subsection{3D Point Cloud Detection on Waymo Open Dataset: Addendum}\label{sec:appendix-wod}

We use the Waymo Open Dataset for 3D Vehicle Object Detection also in its standard setting, with a total 1000 segments of 20s 10Hz videos. We split the 1000 segments into the original 798/202 split.

Our re-implementation of Pointpillar \cite{lang2019pointpillars} is faithful to the topological and threshold hyperparameters of that paper, so we refer the reader to the original work for details. We use 8GPUs and a total batch size of 16, with an Adam optimizer with $(\beta_1, \beta_2) = (0.9, 0.999)$. Our initial learning rate is 0.0015 with a rampup period of 1000 steps. We use a cosine annealing schedule as described in \cite{loshchilov2016sgdr} for a total training regime of 1.28 million steps. 

\cite{lang2019pointpillars} describes eight losses for our bounding boxes: three losses for $(x,y,z)$ localization of the box center, three losses for $(h,\ell,w)$ regression of the box dimensions, one loss for the rotational orientation of the box, and one loss for the binary box class (i.e. vehicle or not). In addition to those losses, we add a ninth loss in the form of a directional classifier; we use a standard cross-entropy loss to predict whether a box faces forwards or backwards within the dataset coordinate system. We use this loss as there is an intrinsic ambiguity in the rotation loss that does not penalize a predicted box for being exactly 180$^o$ rotated with respect to the ground truth. We find that having this additional directional classifier improves performance dramatically on the APH metrics (which penalize heavily for incorrectly oriented boxes).

\begin{figure}[ht]
\begin{subfigure}{.5\textwidth}
  \centering
  \includegraphics[width=0.92\linewidth]{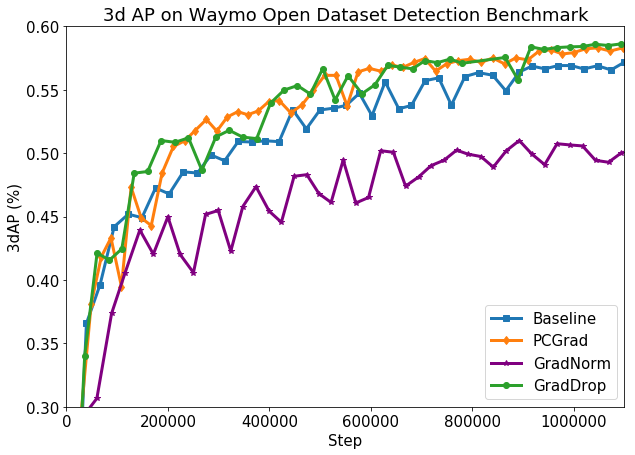}
  \caption{3D AP}
\end{subfigure}
\begin{subfigure}{.5\textwidth}
  \centering
  \includegraphics[width=0.92\linewidth]{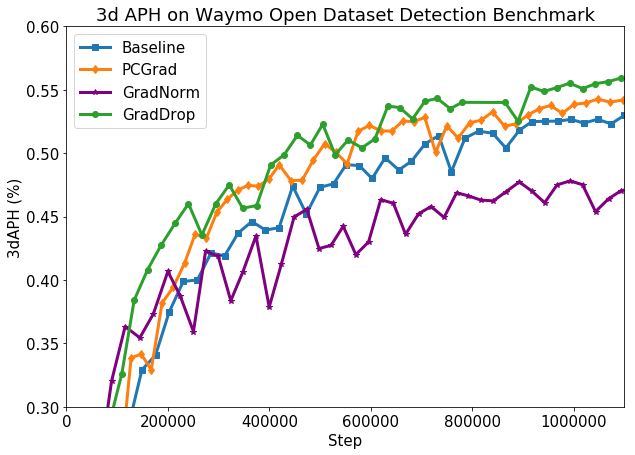}
  \caption{3D APH}
\end{subfigure}
\caption{3D AP and 3D APH metrics for Waymo Open Dataset.}
\label{fig:3ddetection}
\end{figure}

For sake of completeness, we show the accuracy curves for our 3D detection experiments in Figure \ref{fig:3ddetection}. We see that GradDrop produces better accuracy in the 3D detection metrics and this benefit is present throughout most of training. PCGrad \cite{yu2020gradient} also performs well, but falls short of the GradDrop performance. We attribute this differential to the ability of GradDrop to more effectively choose consistent gradient directions, a conclusion also supported by our toy experiments in Section \ref{sec:theory_results}.

As with other experiments the MGDA baseline seems to perform relatively poorly, especially in the 3D metrics. We note that because MGDA seeks the linear combination of gradients that results in the smallest norm, tasks which tend to backpropagate higher gradients will become attenuated by MGDA. GradNorm has a similar effect, but because GradNorm's reference point is the mean norm of all gradients rather than the minimum norm of all possible linear combinations of gradients, the effect is much less acute. Because GradNorm also tends to regress in the 3D metrics, we conclude that the 3D-relevant losses ($z$ localization and box height regression) tend to backpropagate higher gradients, which then has a slight negative interaction with GradNorm and a more severe negative interaction with MGDA.

\begin{table}[b]
  \caption{Object Detection from Point Clouds on the Waymo Open Dataset - Synergy With Other MTL Methods}
  \label{table:appendix-synergy}
  \centering
  \begin{tabular}{lcccc}
    \toprule
    Method     & 2D AP (\%) $\uparrow$    & 2D APH (\%) $\uparrow$& 3D AP (\%)$\uparrow$& 3D APH (\%) $\uparrow$\\
    \midrule
    MGDA \cite{sener2018multi}& 76.8&69.5&20.0&18.3   \\
    MGDA \cite{sener2018multi} + GradDrop& 73.7&65.0&32.3&28.6   \\
    \midrule
    GradNorm \cite{chen2018gradnorm}& 76.9&\textbf{71.7}&51.0&48.2   \\
    GradNorm + GradDrop \cite{chen2018gradnorm}& \textbf{77.3}&\textbf{71.6}&\textbf{55.1}&\textbf{51.5}   \\
    \bottomrule
  \end{tabular}
\end{table}

We also present a more extensive set of results in Table \ref{table:appendix-synergy} for our experiments with synergy between GradDrop and other multitask learning methods, such as GradNorm \cite{chen2018gradnorm} and MGDA \cite{sener2018multi}. In the main paper text we only presented 3D metric results as they were where we saw the most prominent effect, but here we tabulate 2D metrics as well and also present results for MGDA+GradDrop. As described in Section \ref{sec:synergy}, the effect of applying GradDrop atop MGDA is murky; we see a regression in the 2D metrics but an improvement in 3D. As also discussed in Section \ref{sec:synergy}, we attribute this effect to MGDA not behaving properly when its input gradients have the same sign at every position. However, when applied with GradNorm we see that GradDrop significantly improves the 3D metrics while it does approximately as well if not slightly better in the 2D metrics. This result is important, as GradNorm generally provides a moderate boost already in the 2D metrics over the baseline model. The ability of GradDrop to improve the 3D metrics while maintaining the GradNorm advantage in 2D is encouraging. 

\end{document}